\documentclass{article}
\usepackage{ijcai22}
\usepackage{times}

\usepackage{soul}
\usepackage{url}
\usepackage[hidelinks]{hyperref}
\usepackage[utf8]{inputenc}
\usepackage[small]{caption}
\usepackage{graphicx}
\usepackage{amsmath}
\usepackage{amssymb}
\usepackage{xspace}
\usepackage{xcolor}
\usepackage{bm}
\usepackage{booktabs}
\usepackage{amsfonts}
\usepackage{enumitem}
\setenumerate[1]{itemsep=0pt,partopsep=0pt,parsep=\parskip,topsep=2pt}
\setitemize[1]{itemsep=0pt,partopsep=0pt,parsep=\parskip,topsep=2pt}
\setdescription{itemsep=0pt,partopsep=0pt,parsep=\parskip,topsep=2pt}

\usepackage{booktabs}
\usepackage{array}
\usepackage{multirow}
\usepackage{algpseudocode}  
\usepackage{balance}
\newcommand{\mname}{\texttt{MedFACT}\xspace}

\usepackage{algorithm}

\newtheorem{assumption}{Assumption}

\newcommand\blfootnote[1]{%
  \begingroup
  \renewcommand\thefootnote{}\footnote{#1}%
  \addtocounter{footnote}{-1}%
  \endgroup
}

\title{\mname: Modeling Medical Feature Correlations in Patient Health Representation Learning via Feature Clustering}

\author{
Xinyu Ma$^{1,3}$\and
Xu Chu$^2$\and
Yasha Wang$^{1,4}$\footnote{Contact Author}\and
Hailong Yu$^{1,3}$\and
Liantao Ma$^{1,3}$\and
Wen Tang$^{5}$\And
Junfeng Zhao$^{1,3}$
\\
\affiliations
$^1$Key Laboratory of High Confidence Software Technologies, Ministry of Education, Beijing, China\\
$^2$Department of Computer Science, Tsinghua University, Beijing, China\\
$^3$School of Computer Science, Peking University, Beijing, China\\
$^4$National Engineering Research Center of Software Engineering, Peking University, Beijing, China\\
$^5$Division of Nephrology, Peking University Third Hospital, Beijing, China\\
\emails
\{wangyasha, maxinyu\}@pku.edu.cn
}

\begin{document}

\maketitle

\blfootnote{This work has been submitted to the IEEE for possible publication. Copyright may be transferred without notice, after which this version may no longer be accessible.}
\begin{abstract}

In healthcare prediction tasks, it's essential to exploit the correlations between medical features and learn better patient health representations.
Existing methods try to estimate feature correlations only from data, or increase the quality of estimation by introducing task-specific medical knowledge. 
However, such methods either are difficult to estimate the feature correlations due to insufficient training samples, or cannot be generalized to other tasks due to reliance on specific knowledge.
There are medical researches revealing that not all the medical features are strongly correlated. 
Thus, to address the issues, we expect to group up strongly correlated features and learn feature correlations in a group-wise manner to reduce the learning complexity without losing generality.
In this paper, we propose a general patient health representation learning framework \mname.
We estimate correlations via measuring similarity between temporal patterns of medical features with kernel methods, and cluster features with strong correlations into groups.
The feature group is further formulated as a correlation graph, and we employ graph convolutional networks to conduct group-wise feature interactions for better representation learning.
Experiments on two real-world datasets demonstrate the superiority of \mname. The discovered medical findings are also confirmed by literature, providing valuable medical insights and explanations.


\end{abstract}

\section{Introduction}

Nowadays, as electronic medical systems are ubiquitously deployed in hospitals and healthcare centers worldwide, massive electronic health records (EHR) have been rapidly collected. 
EHR is a kind of multivariate time series data, which records various medical features of patients, mainly including static features (e.g., age, gender) and dynamic features (e.g., diagnoses, lab test results).
Recently, deep learning methods have been applied to various health prediction tasks, such as mortality predictions \cite{ma2020concare}, diagnosis predictions \cite{ma2018kame}, etc..
In those tasks, deep learning methods extract patient health representations from massive EHR data, which both effectively and intelligently help doctors estimate patient health status, and conduct targeted treatments to prevent adverse outcomes.

In EHR data, there often exist underlying correlations between medical features.  For instance, \cite{kamal2014estimation} discovers the phenomenon that the simultaneous increase of urea and creatinine can indicate kidney dysfunction, which reveals the strong correlations between those features for patients with chronic kidney disease. 
Therefore, it's of great significance to exploit the inherent correlations between medical features and learn a better patient health representation for downstream prediction tasks.

\begin{figure}[]
\centering
\includegraphics[scale = 0.26]{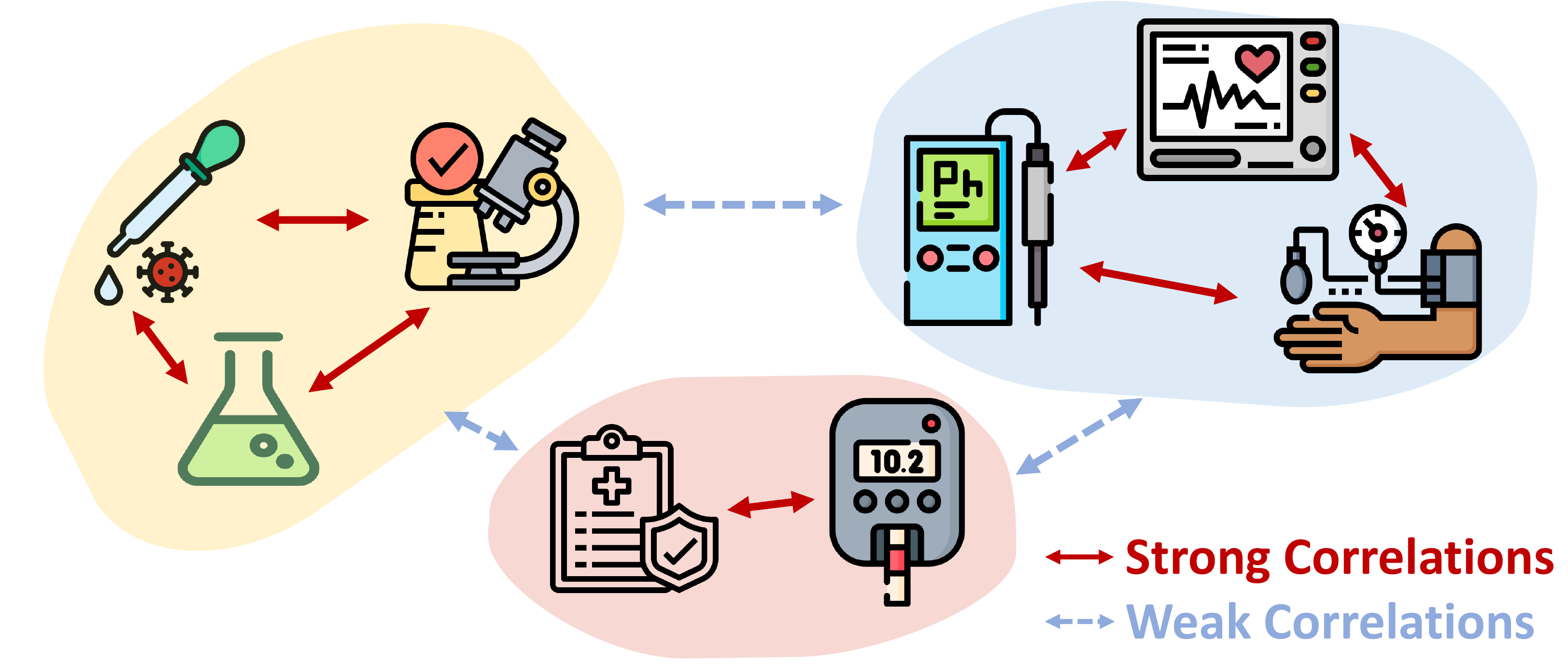}
\caption{Our insight: Medical features are correlated in different ways, and they can be grouped up according to their correlations.}
\label{fig:insight}
\end{figure}

Some existing works try to learn feature correlations with sophisticated deep learning architectures directly from EHR data \cite{choi2016retain,song2018attend,ma2020concare}.
Those networks are designed with highly-parameterized dense layers to deal with high-dimensional medical features and formulate enormous hypothesis spaces.
However, there are often abundant medical features but limited patient samples in EHR data \cite{zhang2019metapred}.
Therefore, it's difficult for deep models to directly learn the correlations between massive features from insufficient patient samples.
Some other works attempt to mitigate this problem by incorporating external prior medical knowledge in deep models \cite{choi2017gram,ma2018kame,lu2021collaborative}.
Those methods utilize various kinds of medical knowledge (e.g., hierarchy of disease code, medical literature) as priors to constrain the hypothesis space of deep networks, which can reduce learning complexity with instructions from prior knowledge.
However, those methods are strongly dependent on precise task-specific knowledge, which is time-consuming and hard to obtain and lack of generalization.
Besides, those methods are not suitable for patient cohorts who suffer from rare diseases or emerging epidemics (e.g., COVID-19), for human experts are lack of knowledge and experience for those diseases.

To tackle the shortcomings above, we expect an approach to constrain learning complexity in deep models while ensuring strong generalization ability at the same time.
Medical researches have revealed that the medical features are not always correlated in the same way. 
Here is a motivating example. \cite{harbarth2001diagnostic} proposes that for sepsis patients, several indicators of cytokines level (e.g., PCT, TNF-$\alpha$) are strongly correlated, while correlations between cytokines and inflammation indicators (e.g., hs-CRP) are weak. 
Therefore, as illustrated in Figure \ref{fig:insight}, we can conclude a prior assumption that medical features can be divided into groups by their correlations, where features in the same group are strongly correlated, and features from different groups are not or weakly correlated.
Then, we can intuitively simplify correlation estimation of all features into several group-wise correlation estimations to reduce the dimension of features in each group and constrain the learning complexity.

Formally, we denote medical features as multi-dimensional stochastic variables $\bm{z_i}$ (i.e. $i$-th medical feature), and our objective, feature correlation learning, is equivalent to approximating the joint probability distribution of features $p(\bm{Z}) = p(\bm{z_1, z_2, ..., z_F})$. We group up features with strong correlations and formulate $K$ subsets of features $\{\bm{G_i}\}_{i=1}^K$, and consider latent features $\bm{L}$ depicting the weak correlations between groups, whose dimension is far less than $\bm{Z}$. 
We present our prior assumption mathematically:
\begin{assumption}
Given latent features $\bm{L}$, features from different groups are \textbf{conditionally independent}:
\begin{equation}
\begin{split}
    p(\bm{Z}) & = p(\bm{G_1},\bm{G_2}, \cdots, \bm{G_K}) \\
    & = p(\bm{G_1}|\bm{L})p(\bm{G_2}|\bm{L}) \cdots p(\bm{G_K}|\bm{L})p(\bm{L})
\end{split}
\end{equation}
\end{assumption}
This provides a solution to simplify the approximation of joint distribution $p(\bm{Z})$, which is reducing $p(\bm{Z})$ to several sub-problems that approximate group-wise feature distributions $p(\bm{G_i}|\bm{L})$ and latent inter-group feature distribution $p(\bm{L})$ with much less features and lower learning complexity. 

However, there remain two practical challenges to realize Assumption 1:

\textbf{1. What is the metric space to measure the correlation between medical features?}
It's necessary to measure feature correlations to divide features into groups reasonably.
For better generalization, we try to measure correlations from data without task-specific external priors.
However, this is still a challenging problem of two entangled perspectives.
1) What is the space to measure? The noise-to-signal ratio is too high in the original representation space, and the coordinate of different features is not well aligned.  A feature mapping guided by the supervised signal could possibly fix this issue.
2) Which metric to use? The support of samples in the latent representation space is a non-Euclidean manifold. This makes it hard to calculate the similarities between samples as well as the correlations between features.

\textbf{2. How to learn group-wise and inter-group feature distributions $p(\bm{G_i}|\bm{L})$ and $p(\bm{L})$?}
It seems straightforward to mask off the feature correlations from different groups to learn group-wise distributions $p(\bm{G_i}|\bm{L})$, such as masked self-attention mechanism.
However, those methods cut off all feature interactions between groups, which cannot solve the latent feature distributions $p(\bm{L})$.
Therefore, it's worth thinking that how to learn $p(\bm{G_i}|\bm{L})$ and $p(\bm{L})$ at the same time.

To address the challenges above, in this paper, we propose a general patient health representation learning framework \mname (\underline{Med}ical \underline{F}e\underline{A}ture \underline{C}lus\underline{T}ering). Our main contributions are summarized as follows:

\begin{itemize}[leftmargin=*]
\item We propose a general health representation learning framework \mname to reduce the learning complexity while capturing correlations between medical features, which does not rely on any task-specific external priors.
\item Specifically, addressing challenge 1, \mname designs a way to group up features according to their underlying correlations exploited from data. We measure feature correlations by computing cohort-wise similarities between temporal patterns of features learned under supervision through a characteristic kernel. The kernel is capable of capturing long-term and short-term dependencies in the signal \cite{gretton2012kernel}. Then we apply spectral clustering algorithm to group up features according to the correlations. 
\item Addressing challenge 2, we formulate the feature group structure as a correlation graph, and graph convolutional network is employed to learn group-wise and latent inter-group feature distributions simultaneously. 
\item Extensive experiments on two real-world datasets demonstrate that \mname significantly outperforms the state-of-the-art methods under various settings. Besides, the discovered findings are confirmed by medical literature, and can also provide valuable medical insights and explanations.
\end{itemize}

\section{Related Works}
There are various works trying to learn feature correlations to generate better health representations for downstream prediction tasks.
Some existing works attempt to learn feature correlations directly from data.
For example, RETAIN \cite{choi2016retain} employs two RNNs to learn different attention weights for visits and medical features.
TimeNet \cite{gupta2018using} designs a shared RNN to encode different features respectively and learn correlations via a linear layer.
TimeLine \cite{bai2018interpretable} and ConCare \cite{ma2020concare} both apply self-attention mechanism to learn feature correlations.
Some other works try to learn feature correlations more accurately by incorporating task-specific medical knowledge.
GRAM \cite{choi2017gram} and KAME \cite{ma2018kame} both incorporate hierarchy of disease codes to enhance learning.
CGL \cite{lu2021collaborative} combines patient personal information with domain knowledge to construct a graph, and uses GCNs to learn better representations.
However, as discussed before, all the methods above are either difficult to learn feature correlations due to insufficient training samples, or lack of generalization due to reliance on task-specific knowledge.
While our method, \mname, can better estimate feature correlations via reducing hypothesis space and learning complexity. Meanwhile, \mname does not rely on any task-specific knowledge, which has better generalization ability.

\begin{figure*}[]
\centering
\includegraphics[scale = 0.53]{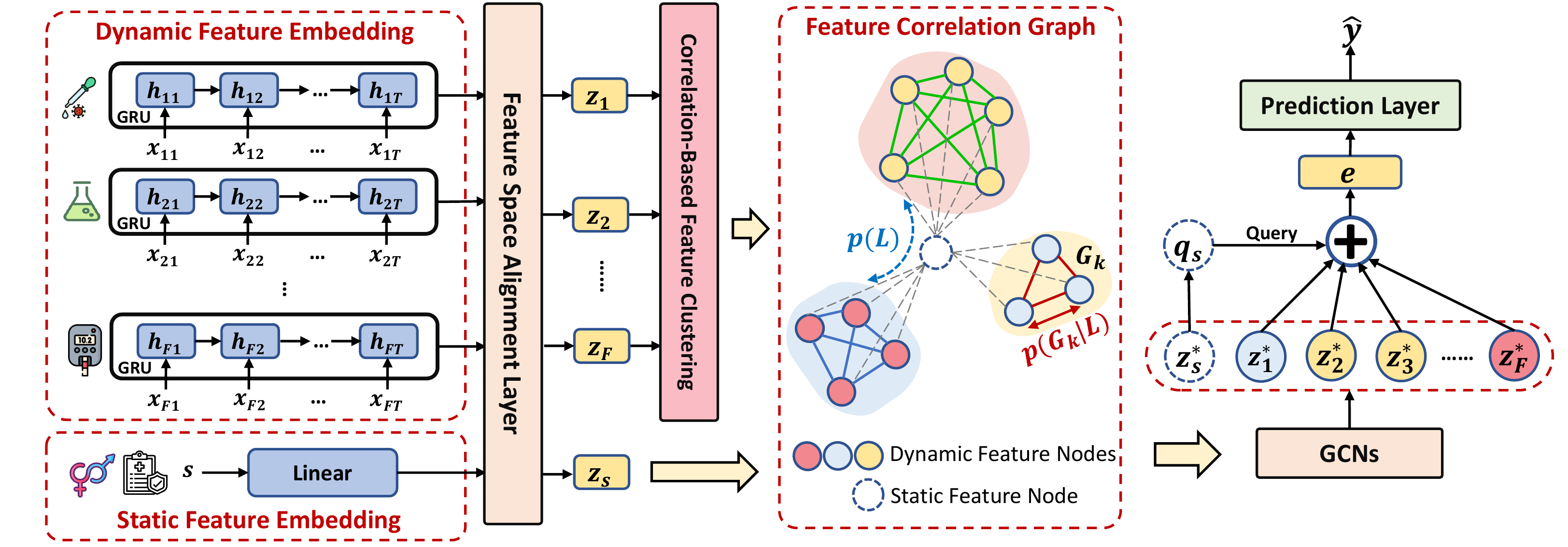}
\caption{The Framework of \mname.}
\label{fig:framework}
\end{figure*}

\section{Problem Formulation}
Electronic Health Records (EHR) data consist of dynamic and static information of patients. For every patient, assuming that there are $F$ dynamic medical features (e.g., lab tests, vital signs) recorded at every clinical visit $t$, the visit can be recorded as a vector $\bm{x_t} \in \mathbb{R}^F$. Supposed that there are $T$ visits, the dynamic information can be formulated as a 2-dimensional matrix $\bm{X} = [\bm{x_1, x_2, ..., x_T}]^\top \in \mathbb{R}^{T \times F}$.
The static information (e.g., demographics) is recorded only once during the whole visits, and can be formulated as a vector $\bm{s} \in \mathbb{R}^S$, where $S$ denotes the number of static features.

In this paper, our predictive objective can be presented as a clinical outcome prediction task. Given the EHR of a patient as inputs, our algorithm attempts to predict the probability of suffering a specific clinical outcome (e.g., mortality), denoted as $y \in \{0, 1\}$. 
We pose this task as a binary classification problem, namely, $\hat{y} = \mname (\bm{X}, \bm{s})$.

\section{Methodology}
Figure \ref{fig:framework} illustrates the general framework of \mname, which comprises the following modules:
\begin{itemize}[leftmargin=*]
    \item \textbf{Feature Embedding Module} learns representations individually for every dynamic and static features.
    \item \textbf{Feature Clustering Module} estimates feature correlations with obtained representations, and the features are clustered into groups according to the estimated correlations.
    \item \textbf{Feature Correlation Graph Construction Module} constructs an edge-weighted correlation graph based on the feature groups and their correlations.
    \item \textbf{Graph-based Feature Interaction Module} employs GCN to learn group-wise and latent inter-group feature distributions based on the correlation graph.
    \item \textbf{Prediction Module} applies attention mechanism and linear layers to generate a comprehensive health representation and conduct specific prediction tasks.
\end{itemize} 

\subsection{Feature Embedding}
In \mname, we expect to estimate the correlations between medical features, which are indicated by similar developing patterns along visits (i.e., simultaneous rising/descending). 
Therefore, \mname attempts to learn representations and extract temporal patterns for every feature from their sequential visits separately.
To address our issue, we utilize multi-channel Gated Recurrent Units (GRU) to extract temporal patterns for every feature individually. 
Specifically, we employ $F$ different GRUs for $F$ features. Each feature $\bm{x_i}$ can be formulated as a time series (i.e., $\bm{x_i} = [x_{i1}, x_{i2}, ..., x_{iT}]$), and will be fed into the corresponding GRU$_i$ for embedding:
\begin{equation}
    \bm{h_{iT}} = \mathrm{GRU}_i (x_{i1}, x_{i2}, ..., x_{iT}), i = 1,2,...,F
\end{equation}
The static features are also embedded into the hidden space with a linear layer: $\bm{h_s} = \bm{s} \cdot \bm{W_s} $. 
Now that we have obtained representations of all features, yet their representation spaces remain unaligned. Therefore, we map all the embeddings into an aligned representation space via a shared activated linear projection $\bm{W}_{proj}$:
\begin{equation}
\begin{split}
    \bm{z_i} &= \mathrm{ReLU}(\bm{h_{iT}} \cdot  \bm{W}_{proj}  ), i = 1,2,...,F \\
    \bm{z_s} &= \mathrm{ReLU}(\bm{h_{s}} \cdot \bm{W}_{proj}  )
\end{split}
\end{equation}
And the representation matrix $\bm{Z}$ stands for the stacked representations: $\bm{Z} = [\bm{z_1}, \bm{z_2}, ..., \bm{z_F}, \bm{z_s}]^\top$.

\subsection{Feature Clustering}
\subsubsection{Feature Correlation Estimation}
In \mname, we expect to divide features into groups according to their correlations in order to simplify the approximation of $p(\bm{Z})$. 
Thus, it's necessary to estimate correlations between features from a cohort-wise perspective. 
We suggest that the correlations between two features can be implied by similar temporal patterns that widely appear in the patient cohort.
Now that the temporal patterns of each feature $i$ are extracted by the GRU embedded feature representation $\bm{z_i}$ under the supervision of labels, it's natural to consider the cohort-wise similarities of $\bm{z_i}$ as the feature correlations.

In \mname, we employ the characteristic kernel method \cite{gretton2012kernel} to measure the sample-wise similarity of two features in non-Euclidean latent space. Here we select Laplacian kernel (i.e. $k(\bm{x}, \bm{y}) = \exp{(-\frac{\Vert \bm{x} - \bm{y} \Vert_1}{\sigma})}$) as the kernel function. 
We suppose that the cohort-wise similarities can be approximated by the average of sample-wise similarities, and the correlations between two features $i$ and $j$ can be defined as:
\begin{equation}
    r_{ij} = \frac{1}{N} \sum_n^N k(\bm{z_i}^{(n)}, \bm{z_j}^{(n)}).
\end{equation}
Thus the correlations between all features can be formulated as a matrix $\bm{R} =(r_{ij})_{F \times F}$.

\subsubsection{Correlation-based Feature Clustering}
\mname attempts to apply clustering algorithms to group up features based on the learned correlations, and expects to make features in the same group have stronger correlations.
In detail, \mname utilizes K-Means based spectral clustering algorithm \cite{stella2003multiclass} to cluster all $F$ dynamic features into $K$ groups based on the pre-computed affinity matrix $\bm{R}$. We denote the $K$ groups as subsets of features (i.e. $\bm{G_1}, \bm{G_2}, ..., \bm{G_K}$), satisfying:
\begin{equation}
    \bigcup_{i}^{K}\bm{G_i} = \{\bm{z_1},\bm{z_2},...,\bm{z_F}\}; \quad \bm{G_i} \cap \bm{G_j} = \varnothing, \forall i \neq j
\end{equation}
The optimization objective of spectral clustering can be formulated as follows:
\begin{equation}
    \min_{\bm{G_1}, \bm{G_2}, ..., \bm{G_K}} \sum_{k=1}^K \sum_{\bm{z_i}\in \bm{G_k}} \sum_{\bm{z_j} \notin \bm{G_k}} r_{ij},
\end{equation}
which aims to minimize the sum of correlations between the features from different groups. 
Besides, we also assume that all the dynamic features are strongly related with static features, and we make static features belong to every group:
\begin{equation}
    \bm{G_i} :=  \bm{G_i} \cup \{\bm{z_s}\},  i=1,...,K 
\end{equation}

\subsection{Feature Correlation Graph Construction}
We try to model the group structure of medical features in a more plain view.
Intuitively, we convert the feature group structure to a specific correlation graph, where the graph nodes denote medical features, and their connections denote the group-wise correlations.
Concretely, there are $F+1$ nodes in the graph, including $F$ dynamic feature nodes and one static feature node. The weighted edge between nodes denotes the correlations between features. We use an adjacency matrix $\bm{A}$ to represent the graph, whose element $a_{ij} \in [0, 1]$ denotes the correlation weight between feature $i$ and $j$. There are three specific graph construction rules:
\begin{itemize}[leftmargin=*]
    \item All the dynamic features from the same feature group $k$ are connected with each other, forming a fully-connected subgraph. Specifically, for any dynamic feature pair $\bm{z_i}, \bm{z_j} \in \bm{G_k}, i \neq j$, there's an edge between them, and $a_{ij}=r_{ij}$.
    \item For the static feature node, we have supposed that all the features are strongly related with it. Specifically, there's an edge between every dynamic feature $\bm{z_i}$ and the static feature, and we set the correlation weight $a_{i(F+1)}=1$.
    \item All the features are self-related, which means there's a self-connection on every feature $\bm{z_i}$ weighted $a_{ii}=1$.
\end{itemize}
The topology of the constructed feature correlation graph is illustrated in Figure \ref{fig:framework}, where different colors of feature nodes denote different groups, and edges denote the correlations.

\subsection{Graph-based Feature Interaction}
According to Assumption 1, approximation of $p(\bm{Z})$ is reduced to sub-problems solving group-wise feature distributions $p(\bm{G_i}|\bm{L})$ and latent inter-group feature distribution $p(\bm{L})$.
Inspired by Graph Convolutional Network (GCN) \cite{kipf2016semi}, \mname tries to interact information from neighbor nodes based on the correlation graph, and apply 2 GCN layers to solve $p(\bm{G_i}|\bm{L})$ and $p(\bm{L})$ respectively.


Specifically, a GCN layer $g_l(\cdot)$ conducts feature transformation with a parameter matrix $\bm{W}_l$ and further interacts features from all neighbor nodes based on adjacency matrix $\bm{A}$:
\begin{equation}
    g_l(\bm{Z}) = \mathrm{ReLU}(\bm{A} \bm{Z} \bm{W}_l),
\end{equation}
where the footnote $l$ denotes the $l$-th layer of GCN. Due to the special property of the correlation graph, the first layer of GCN $g_1(\bm{Z})$ only conduct interactions between feature nodes in the same group, which learns the group-wise conditional joint distributions $p(\bm{G_i}|\bm{L})$. 
After $g_1(\bm{Z})$, the static feature node combines information from all features. 
Therefore, in the second layer of GCN $g_2(\bm{Z})$, all the feature nodes can extract information from any other features via static node. That means besides group-wise correlations, $g_2(\bm{Z})$ can also learn the latent feature distribution $p(\bm{L})$ that depicts the weak inter-group correlations.

After two layers of GCN, we have solved all the reduced sub-problems and is able to approximate the joint distribution $p(\bm{Z})$.
The features after interaction are denoted as a matrix $\bm{Z}^* =  [\bm{z_1^*}, \bm{z_2^*}, ..., \bm{z_F^*}, \bm{z_s^*}]^\top$, where:
\begin{equation}
    \bm{Z}^* = g_2( g_1(\bm{Z})) = \mathrm{ReLU}(\bm{A} \  \mathrm{ReLU}(\bm{A} \bm{Z} \bm{W}_1) \  \bm{W}_2).
\end{equation}

\subsection{Prediction Layers}
Finally, a comprehensive health representation of a patient is expected to perform the personalized prediction.
Here we introduce an attention mechanism to summarize information from the representations $\bm{Z}^*$.
Concretely, we use the representation of static features $\bm{z_s^*}$ to obtain the query $\bm{q_s}$, while the keys $\bm{k_i}$ and values $\bm{v_i}$ are obtained by $\bm{z_i^*}$:
\begin{equation}
\begin{split}
    \bm{q_s} & = \bm{z_s^*} \cdot \bm{W_q}; \\
    \bm{k_i} = \bm{z_i^*} \cdot \bm{W_k}, \ \bm{v_i} & = \bm{z_i^*} \cdot \bm{W_v}, \ i=1,2,..,F,s
\end{split}
\end{equation}
where $\bm{W_q}$, $\bm{W_k}$, and $\bm{W_v}$ are projection matrices, and the attention weights $\alpha_i$ are calculated as:
\begin{equation}
\begin{split}
    \tau_i & = \mathrm{tanh}\bm{(q_s}^\top \bm{k_i}), \\
    \alpha_1, ..., \alpha_F, \alpha_s & = \mathrm{Softmax}(\tau_1, ..., \tau_F, \tau_s).
\end{split}
\end{equation}
The comprehensive health representation $\bm{e}$ is obtained by weighted sum of values $\bm{v_i}$, and we use a linear layer to conduct the final prediction task based on $\bm{e}$:
\begin{equation}
    \bm{e} = \sum_{i=1}^F \alpha_i \bm{v_i} + \alpha_s \bm{v_s}, \  \hat{y} = \mathrm{Sigmoid}(\bm{e} \cdot \bm{W}_{pred}).
\end{equation}
Finally, the cross-entropy loss is applied as the loss function:
\begin{equation}
    \mathcal{L} = - y \log(\hat{y}) - (1-y) \log(1-\hat{y}),
\end{equation}
where $\hat{y} \in [0,1]$ is the predicted probability and $y$ is the ground truth.
We present the detailed algorithm of \mname in Algorithm \ref{alg:full_model}. It's worth mention that in order to ensure stability in training, we don't always update the correlation graph at the end of every training epoch. After certain epochs (i.e. CLUSTER\_EPOCHS), the correlation graph is fixed and no longer updated.
In experiments, we empirically set CLUSTER\_EPOCHS to 20\% of the total training epochs.

\begin{algorithm}
\caption{\mname $(\bm{X}, \bm{s})$}
\label{alg:full_model}
\begin{algorithmic}[1]

\State \textbf{Initialization}: Set all elements in correlation matrix $\bm{R}$ to 1. Randomly initiate $K$ clusters and construct the correlation graph following Section 4.3. 
\While {$E$ in epochs}:
    \For {$B$ in mini-batches}:
        \State Compute batch loss function $\mathcal{L}$ 
        \State Update parameters of $\mname$ by optimizing $\mathcal{L}$ 
    \EndFor
    \If {$E <$ CLUSTER\_EPOCHS}:
        \For {$i$ in 1,2,...,$F$}:
            \For {$j$ in 1,2,...,$F$}:
                \State {Calculate  $\bm{R}[i,j] = \frac{1}{N} \sum_n^N k(\bm{z_i}^{(n)}, \bm{z_j}^{(n)})$}
            \EndFor
        \EndFor
        \State {Cluster features to $K$ groups $\{\bm{G_i}\}_{i=1}^K$ according to the correlations $\bm{R}$, and update correlation graph $\bm{A}$.}
    \EndIf
\EndWhile
\end{algorithmic}
\end{algorithm}

\section{Experiments}
\subsection{Dataset Descriptions}
\paragraph{CKD Dataset} We conduct mortality prediction task on a real-world chronic kidney disease (i.e. CKD) dataset, including CKD patients who received therapy from January 1, 2006, to March 1, 2018, in a real-world hospital.\footnote{This study was approved by the Research Ethical Committee.} 
The mortality prediction task is formulated as a binary classification task, predicting whether the patient dies unfortunately within a year after the last visit.
The cleaned dataset consists of 662 patients with 17 dynamic features (e.g., glucose) and 4 static features (e.g., gender).
The detailed statistics are presented in Appendix.
Due to the scarcity of CKD data, 5-fold cross-validation experiments are performed.

\paragraph{Cardiology Dataset} Another dataset we use is an open-source PhysioNet cardiology dataset \cite{reyna2019early}, which is collected from three geographically distinct U.S. hospitals over the past decade.
The patients in the dataset are binary labeled by Sepsis-3 clinical criteria, and we conduct the sepsis prediction on it.
This dataset is highly imbalanced, with only 7.26\% of samples labeled positive.
The cleaned dataset consists of 40,336 patients with 34 dynamic features and 5 static features, and the detailed statistics are presented in Appendix.
The dataset is randomly divided into the training, validation, and testing set with a proportion of 8:1:1.

\subsection{Experimental Setups}
We implement our methods with PyTorch v1.7.1 and conduct experiments on a machine equipped with GPU: Nvidia Quadro RTX 8000. 
While training models, Adam optimizer is employed with learning rate set to 1e-3.
To fairly compare different approaches, the hyperparameters of the models are fine-tuned by grid search on training sets.
Specifically for Cardiology dataset, the number of clusters $K$ is set to 6, and for CKD dataset, we set $K$ to 4.

\paragraph{Evalutation Metrics} 
We assess the performance with three evaluation metrics: area under the receiver operating characteristic curve (AUROC), area under the precision-recall curve (AUPRC) and the minimum of precision and sensitivity (Min(P+,Se)). 
Among those metrics, AUPRC is a more informative and primary metric when dealing with a highly imbalanced dataset like ours. \cite{davis2006relationship}

\paragraph{Baseline Methods} 
We select several state-of-the-art methods as our baselines. Baselines incorporating external medical knowledge are not included:
\begin{itemize}[leftmargin=*]
    \item GRU$_\alpha$ is the naive GRU with attention mechanism.
    \item RETAIN \cite{choi2016retain} (NeurIPS) applies different attention weights on different visits and medical features.
    \item T-LSTM \cite{baytas2017patient} (SIGKDD) tackles time intervals by introducing a time decay mechanism in LSTM.
    \item TimeNet$_*$ \cite{gupta2018using} (IJCAI) designs a shared encoder to embed all features respectively and learn correlations via linear layers. For fair comparison, pretraining is not conducted here.
    \item StageNet \cite{gao2020stagenet} (WWW) integrates patient disease stage development to learn feature correlations.
    \item ConCare \cite{ma2020concare} (AAAI) embeds each medical features individually, and employs self-attention mechanism to model and interpret feature correlations.
\end{itemize}

We also perform the following ablation studies:
\begin{itemize}[leftmargin=*]
    \item \mname$_{cor-}$ doesn't consider any difference on feature correlations. The correlation graph is fully connected with all edge weights set to 1.
    \item \mname$_{clu-}$ doesn't cluster features into groups. The correlation graph is fully connected, and the adjacency matrix $\bm{A}$ equals to the correlation matrix $\bm{R}$.
\end{itemize}

\begin{table*}[]
\centering
\caption{Experimental Results for Tasks on Cardiology \& CKD Datasets}
\label{tab:results}
\begin{tabular}{ccccccc}
\toprule

 & \multicolumn{3}{c}{Mortality Prediction on CKD Dataset} & \multicolumn{3}{c}{Sepsis Prediction on Cardiology Dataset}  \\
Methods  & AUPRC & AUROC  & Min(P+,Se) & AUPRC & AUROC  & Min(P+,Se) \\ 
\midrule
GRU$_\alpha$ & 0.6790(.061) & 0.7849(.041) & 0.6397(.048) &  0.6356(.030)  & 0.9222(.009) & 0.6389(.024) \\

RETAIN & 0.6938(.053) & 0.8018(.034) & 0.6744(.027) & 0.7310(.023) & 0.9386(.008) & 0.6681(.023)  \\

T-LSTM   & 0.6976(.056)  & 0.8012(.033) & 0.6670(.049) & 0.6866(.024) & 0.9258(.009) & 0.6131(.024) \\

TimeNet$_*$  & 0.7353(.038) & 0.8219(.024) & 0.6758(.040)& 0.7829(.020) & 0.9472(.007) & 0.7004(.022)  \\

StageNet & 0.7301(.029) & 0.8120(.030) & 0.6729(.057) & 0.7170(.024) & 0.9421(.007) & 0.6497(.023)  \\

ConCare  & 0.7295(.079)  & 0.8226(.039) & 0.6727(.054)  & 0.7740(.021) & 0.9512(.006) & 0.7037(.021) \\

\hline

\mname$_{cor-}$ & 0.7469(.048) & 0.8273(.029) & 0.6786(.034) & 0.7795(.021) & 0.9522(.006) & 0.7176(.022) \\

\mname$_{clu-}$ & 0.7484(.059) & 0.8323(.034) & 0.6881(.034) & 0.7823(.021) & 0.9552(.006) & 0.7146(.021) \\

\hline

\mname    & \textbf{0.7653(.039)} & \textbf{0.8357(.027)} & \textbf{0.6921(.046)}  & \textbf{0.8047(.020)} & \textbf{0.9556(.006)} & \textbf{0.7311(.021)}\\

\bottomrule
\end{tabular}
\end{table*}

\subsection{Experimental Results}
Table \ref{tab:results} shows the performance of \mname and baselines on the two datasets.
The value in ($\cdot$) denotes the standard deviation of 1000-times bootstrapping and 5-fold cross-validation for Cardiology and CKD dataset, respectively.
As presented in Table \ref{tab:results}, we observe that \mname significantly outperforms all other baselines on both datasets.
This is mainly because it's difficult for all the baselines (including ablation studies) to directly learn global feature correlations $p(\bm{Z})$ from data, while \mname reduces it to several sub-problems, which lowers the learning complexity and brings improvements on performance.
Especially for datasets with fewer samples, $p(\bm{Z})$ is much harder to learn, and \mname can better mitigate this problem and reach higher performance. 
The conclusion gets proved on CKD dataset with much fewer patient samples. The relative performance boost reaches 4.08\% on AUPRC and 1.59\% on AUROC compared with the best baseline, which is much larger than that on Cardiology dataset (2.78\% on AUPRC and 0.46\% on AUROC). 
Besides, comparing the two ablation studies, \mname$_{clu-}$ shows better performance than \mname$_{cor-}$, indicating that it's also effective to incorporate correlation differences in the model.

\subsection{Analysis}
\begin{figure}[]
\centering
\includegraphics[scale = 0.43]{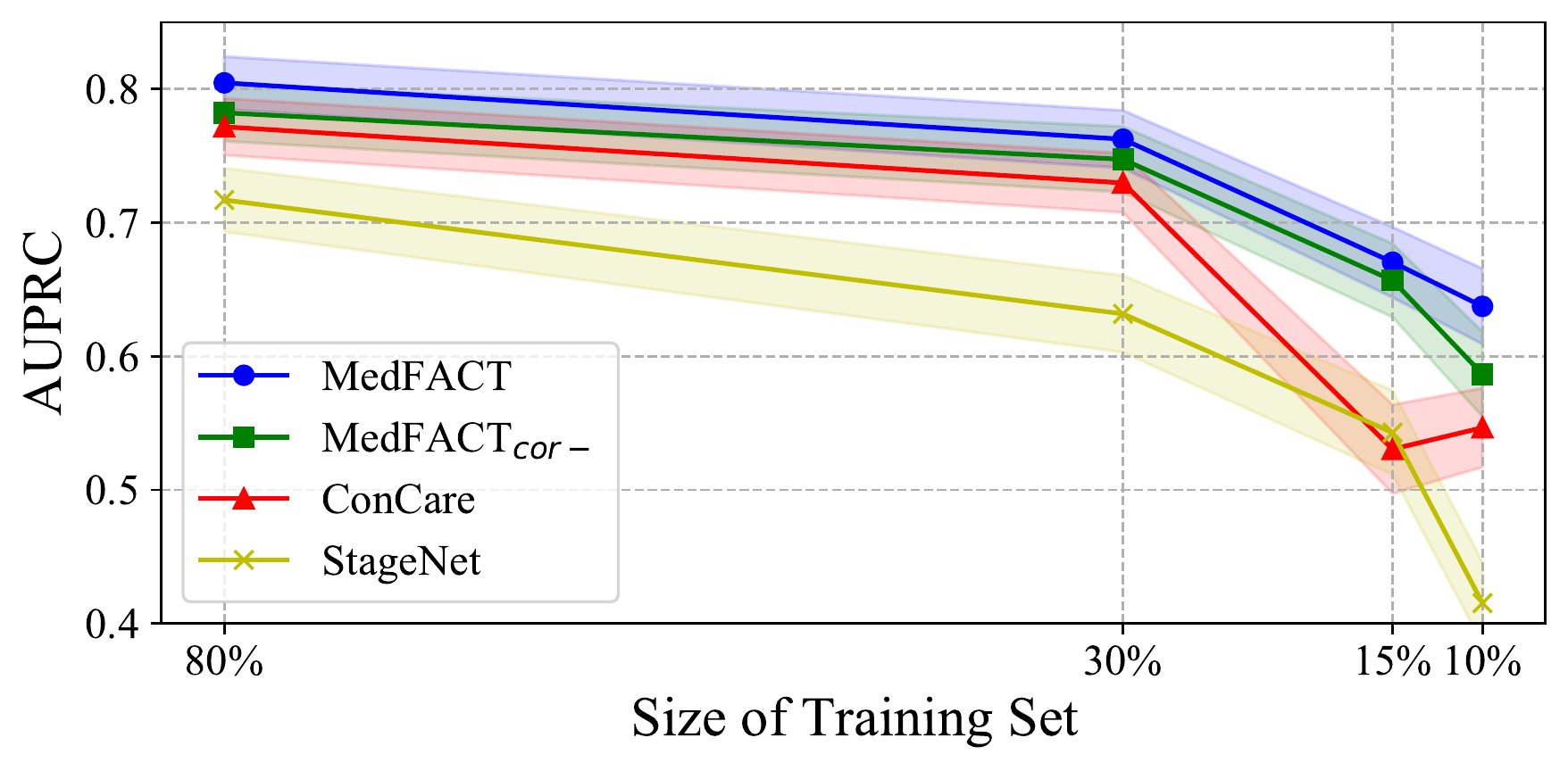}
\caption{Performance comparison of \mname and several baselines with different amount of training data on Cardiology dataset. }
\label{fig:vary_datasize}
\end{figure}

\begin{figure}[]
\centering
\includegraphics[scale = 0.43]{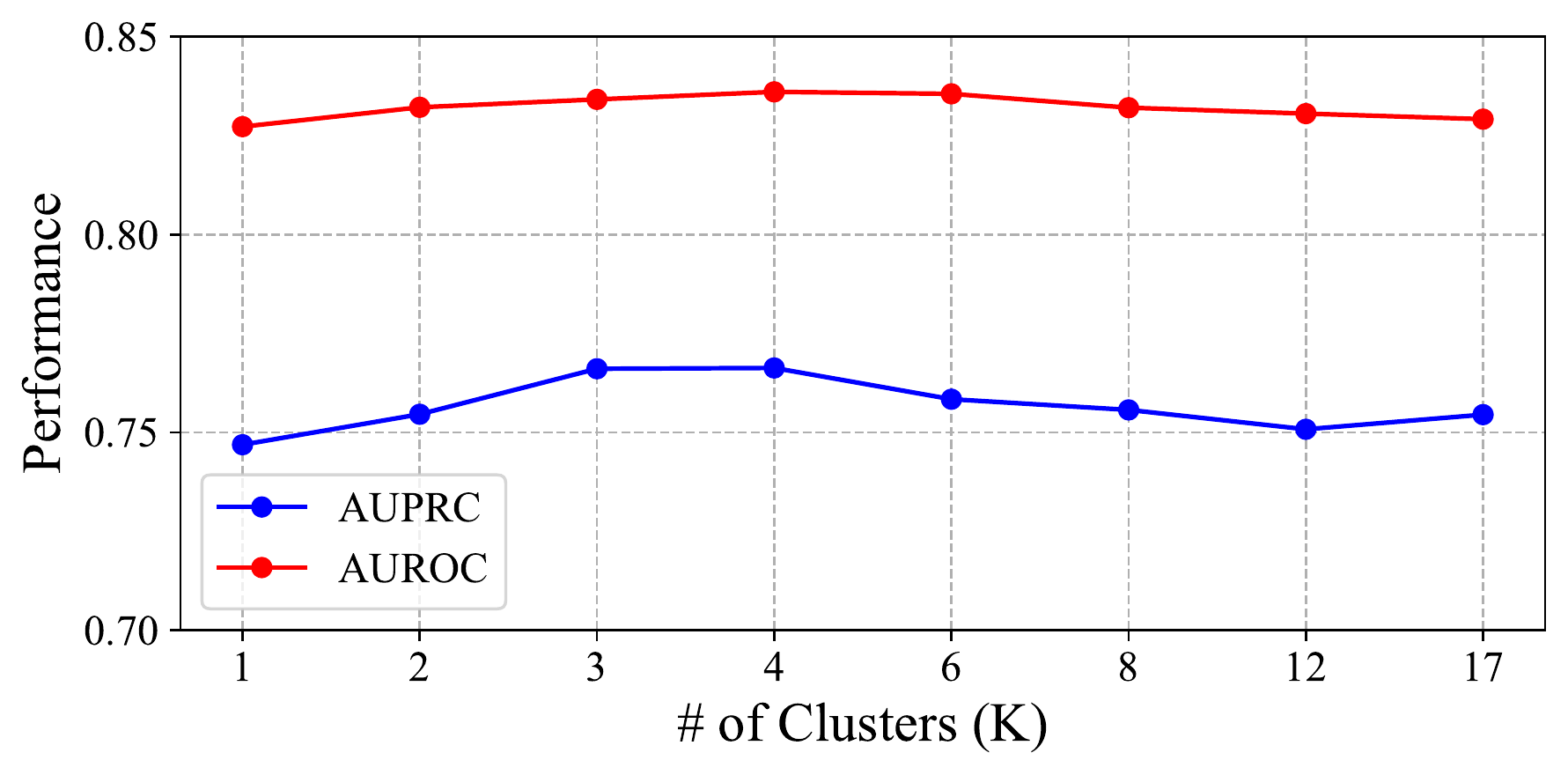}
\caption{Performance for different $K$ settings on CKD dataset.}
\label{fig:vary_clu_num_perf}
\end{figure}

\paragraph{Varying the Data Size}
We try to evaluate the robustness of \mname against insufficient training samples.
Here we reduce the training set of Cardiology dataset from 80\% to 30\%/15\%/10\% of the whole dataset to simulate the scenario of data insufficiency, and the test set is fixed for a fair comparison.
We conduct experiments on those different settings, and the AUPRC ($\pm$ std.) is plotted in Figure \ref{fig:vary_datasize}. 
As is shown in Figure \ref{fig:vary_datasize}, \mname consistently outperforms all selected baselines under all settings.
Furthermore, as the size of training set shrinks, the performances of other methods decrease more sharply than ours, leading to a larger performance gap.
Even when we adopt only 10\% of data for training, \mname still reaches an AUPRC of 0.6374, while \mname$_{cor-}$ and ConCare drop to 0.5866 and 0.5466, showing 8.66\% and 16.6\% relative improvement, respectively.
Those results indicate that \mname is more tolerant of data insufficiency, and demonstrate the robustness of our method.

\begin{figure}[]
\centering
\includegraphics[scale = 0.23]{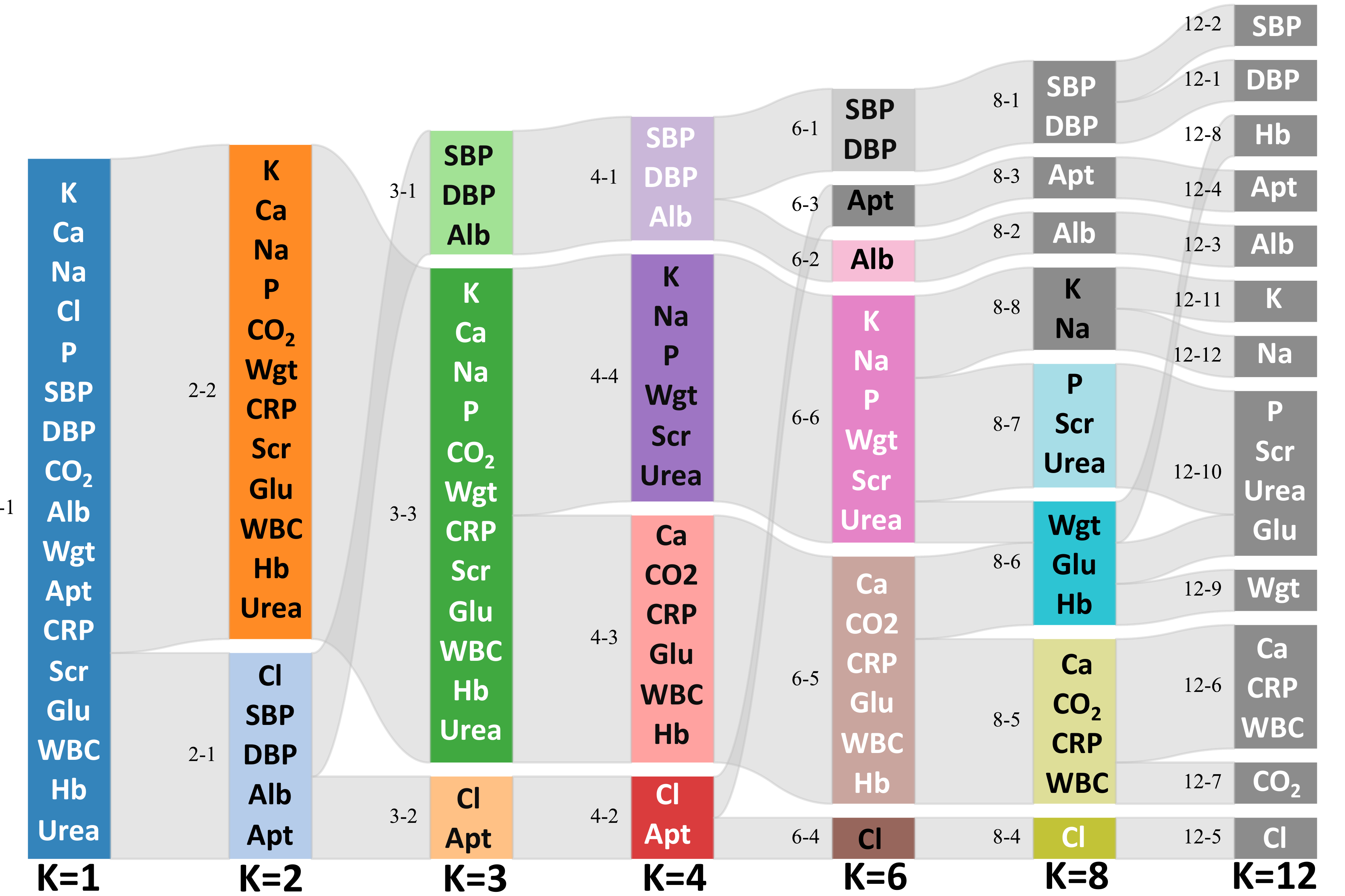}
\caption{Evolution of feature clusters on CKD dataset as the number of clusters $K$ increases. Clusters gradually split up as $K$ increases and formulate a hierarchical tree structure.}
\label{fig:vary_clu_num_sankey}
\end{figure}

\paragraph{Varying the Number of Clusters $K$} 
We try to observe the performance of \mname under different settings of $K$ on CKD dataset.
As is shown in Figure \ref{fig:vary_clu_num_perf}, the performance peaks at $K=4$. Selecting a $K$ that is too large or too small can lead to approximately 1\% of performance decay.
That's because each cluster still contains lots of features if $K$ is too small, making it hard to learn the group-wise correlations (i.e., $p(\bm{G_i}|\bm{L})$).
Meanwhile, if $K$ is too large, there can be massive clusters but scarce features in each cluster, making the inter-group correlations (i.e., $p(\bm{L})$) hard to estimate.

Besides, it's also interesting to analyze how feature clusters evolve as $K$ increases. 
Figure \ref{fig:vary_clu_num_sankey} illustrates that the feature clusters gradually split up as $K$ increases, and there are seldom features switched to another cluster when $K$ changes, demonstrating the stability of our cluster results.
The gradual split-up of clusters formulate a hierarchical structure of medical features, which can be concluded as medical knowledge obtained from data.
Furthermore, the feature correlations in clusters are confirmed by medical literature.
For example, \cite{jiang2020u} shows positive correlations between albumin, SBP, and DBP in CKD patients, which matches cluster 4-1. 
This cluster is further split into two parts, one of which (i.e. cluster 6-1) is SBP and DBP, the blood pressure indicators.
Another example is cluster 4-4. \cite{ellison2017treatment} claims that electrolyte disorder is common in CKD patients, which can further lead to tissue edema and an increase of body weight. This exactly matches cluster 4-4, where K, Na, P are all features depicting electrolyte balance, and weight(Wgt) is also included.
We also find that serum albumin(Alb), chloride(Cl), and  appetite(Apt) are individually clustered after $K=6$.
Medical researches \cite{kubota2020prognostic,grove2018self,menon2005c} reveal that those features are all independent predictors of mortality in CKD, indicating that our model learns a more distinguishable representation space for those features.

\section{Conclusion}
In this paper, we propose a health representation learning framework \mname to reduce learning complexity while modeling medical feature correlations without external task-specific knowledge.
\mname groups up features with strong correlations, and reduces global feature correlation estimation to several sub-problems estimating group-wise and inter-group correlations. Specifically, \mname designs a novel metric to cluster features, and construct a graph to learn feature correlations via GCNs. 
\mname demonstrates significant performance improvements, provides medical knowledge and discovers reasonable feature clusters that match medical literature. We hope \mname can provide valuable medical insights and help physicians better diagnose.

\newpage
\bibliographystyle{named}
\bibliography{ijcai22}

\appendix

\end{document}


\appendix



\section{Dataset Statistics}
In this section, we introduce the dataset we use (i.e. CKD dataset and Cardiology dataset) in experiments in detail.
Specifically, CKD dataset is recorded with chronic patients, and their visits are generated monthly.
Cardiology dataset only consists of ICU patients, and the visits are recorded hourly.
We present the detailed statistics in Table \ref{tab:dataset}, and all the dynamic and static medical features are listed in Table \ref{tab:features}.

\begin{table}
  \centering
  \caption{Statistics of Datasets}
  \label{tab:dataset}
  \begin{tabular}{lcc}
    \toprule
   Dataset & CKD & Cardiology  \\
    \midrule
    \# of patients & 662 & 40,336 \\
    \# of visits  & 13,108 & 1,552,210 \\
    Avg. visits / patient & 19.95& 38.48 \\
    Max. visits / patient & 69 & 336 \\
    Min. visits / patient & 1 & 8\\
    \# of dynamic features & 17 & 34 \\
    \# of static features & 4 & 5  \\
    \% of positive labels & 38.97\% & 7.26\% \\
  \bottomrule
\end{tabular}
\end{table}

\begin{table}[]
  \centering
  \caption{Features Recorded in CKD and Cardiology Dataset}
  \label{tab:features}
  \begin{tabular}{c|c|cc}
  \toprule
   & CKD Features & \multicolumn{2}{c}{Cardiology Features} \\
  \hline
    \multirow{17}{*}{Dynamic} & Systolic BP & Heart Rate & O$_2$Sat \\
    & Diastolic BP& Temperature & Systolic BP\\
    & Urea & Diastolic BP & MAP \\
    & Calcium & Resp. Rate & EtCO$_2$\\
    & Chloride & BaseExcess & HCO$_3$\\
    & Creatinine & FiO$_2$ & pH\\
    & Glucose & PaCO$_2$ & SaO$_2$\\
    & Phosphate & AST & BUN \\
    & Potassium & Alkalinephos & Calcium\\
    & Hemoglobin & Chloride & Creatinine\\
    & WBC Count& Dir. Bilirubin & Glucose\\
    & Sodium & Lactate & Magnesium \\
    & CO$_2$CP & Phosphate & Potassium \\
    & Albumin & Tot. Bilirubin & TroponinI\\
    & Weight & Hematocrit & Hemoglobin\\
    & Appetite & PTT & WBC Count\\
    & hs-CRP & Fibrinogen & Platelets \\
  \hline
    \multirow{4}{*}{Static} & Age & Age & Gender \\
    & Gender & MICU & SICU\\
    & Height & HospAdmTime & \\
    & Diabetes & &\\
  \bottomrule
\end{tabular}
\end{table}

\section{Varying Number of Clusters $K$ on Cardiology Dataset}

\begin{figure}[]
\centering
\includegraphics[scale = 0.43]{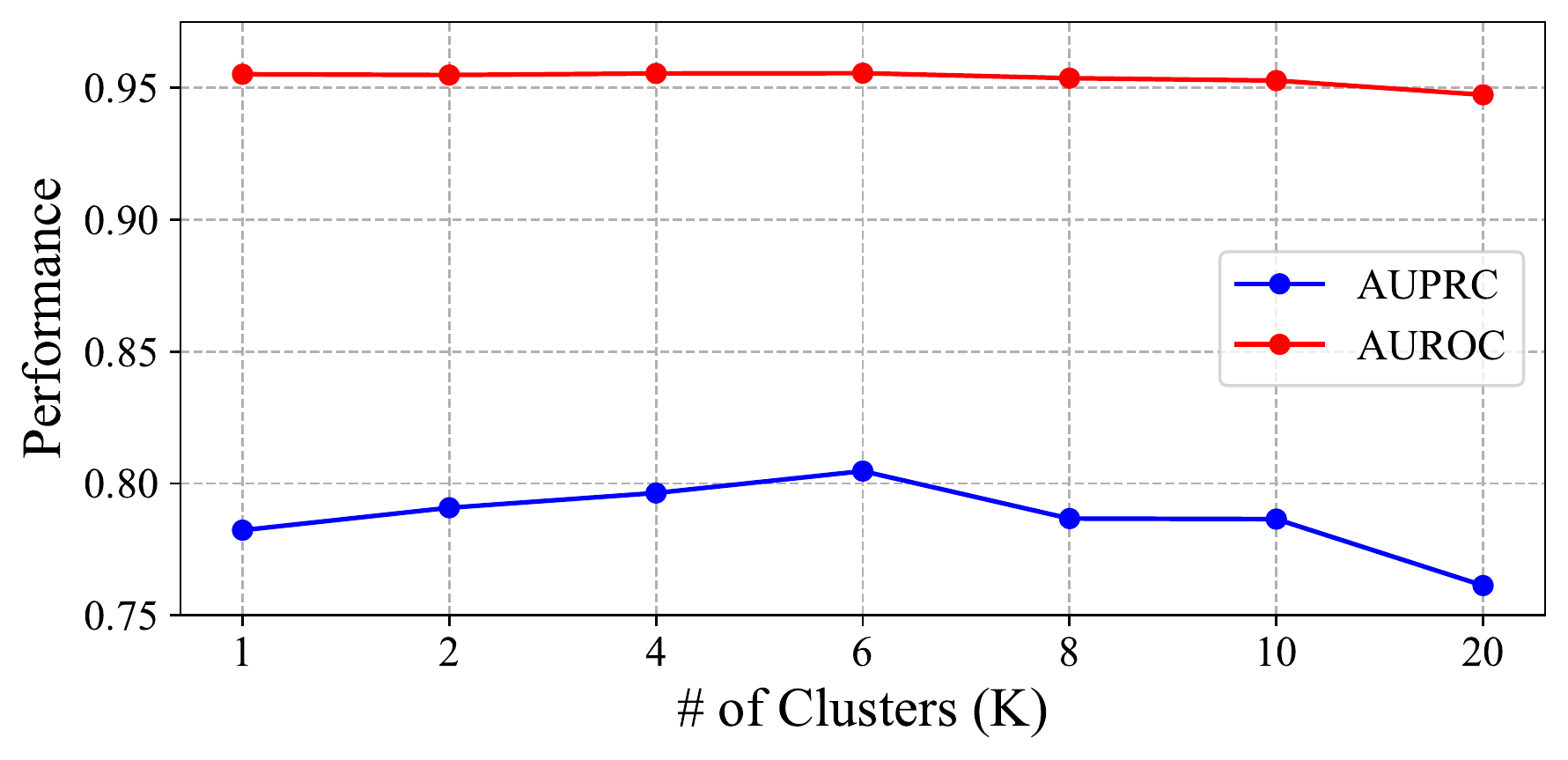}
\caption{AUROC and AUPRC performance for different $K$ settings on Cardiology dataset.}
\label{fig:vary_clu_num_challenge}
\end{figure}

We have already studied how number of clusters $K$ affects the performance of \mname on CKD dataset in our main text. Here we also provide similar analysis on Cardiology dataset.
Figure \ref{fig:vary_clu_num_challenge} shows the performance of \mname under different settings of $K$.
The results indicate the same conclusions as CKD that selecting a $K$ too large or too small will make it challenging to learn group-wise or inter-group feature distributions $p(\bm{G_i|Z})$ or $p(\bm{Z})$ and lead to performance decay.
Here we present a possible intuitive approach to select a better $K$ which may balance the learning complexity of $p(\bm{G_i|Z})$ and $p(\bm{Z})$.
Assume that the learning complexity of $p(\bm{G_i|Z})$ is a function of the number of features in each group (i.e. $\phi(F/K)$), and the learning complexity of $p(\bm{Z})$ is a function of the number of clusters (i.e. $\phi(K)$). 
Then the balance of learning complexity is achieved when $F/K=K$, which is $K=\sqrt{F}$.
This provides an intuitive approximation of $K$ for better performance.
As a result, $K$ is set to $6\approx\sqrt{34}$ for Cardiology dataset, and $K$ is $4\approx\sqrt{17}$ for CKD dataset, and the performances reach their peak at those settings.

